\documentclass[runningheads]{llncs}
\usepackage{amsmath}
\usepackage{amssymb}
\usepackage[T1]{fontenc}

\usepackage{graphicx}
\begin{document}
%
\title{Memory Sequence Length of Data Sampling Impacts the Adaptation of Meta-Reinforcement Learning Agents}
\titlerunning{Memory Sequence Length Impacts the Adaptation of Meta-RL Agents}



%
%
\author{Menglong Zhang\inst{1}\orcidID{0009-0000-0304-3178} \and
Fuyuan Qian\inst{1}\orcidID{0009-0002-2560-7872} \and
Quanying Liu\inst{1}\orcidID{0000-0002-2501-7656}}


%
\authorrunning{Zhang et al. 2024}
%
\institute{$^1$ Department of Biomedical Engineering, Southern University of Science and Technology, Shenzhen, 518055, China\\ 
Corresponding to \email{liuqy@sustech.edu.cn} (Q.Liu)\\
}

\maketitle              
\begin{abstract}
Fast adaptation to new tasks is extremely important for embodied agents in the real world. Meta-reinforcement learning (meta-RL) has emerged as an effective method to enable fast adaptation in unknown environments. Compared to on-policy meta-RL algorithms, off-policy algorithms rely heavily on efficient data sampling strategies to extract and represent the historical trajectories. However, little is known about how different data sampling methods impact the ability of meta-RL agents to represent unknown environments. Here, we investigate the impact of data sampling strategies on the exploration and adaptability of meta-RL agents. Specifically, we conducted experiments with two types of off-policy meta-RL algorithms based on Thompson sampling and Bayes-optimality theories in continuous control tasks within the MuJoCo environment and sparse reward navigation tasks. Our analysis revealed the long-memory and short-memory sequence sampling strategies affect the representation and adaptive capabilities of meta-RL agents. We found that the algorithm based on Bayes-optimality theory exhibited more robust and better adaptability than the algorithm based on Thompson sampling, highlighting the importance of appropriate data sampling strategies for the agent's representation of an unknown environment, especially in the case of sparse rewards.

\end{abstract}
\section{Introduction}

The realization of embodied intelligence relies on an agent's ability to fast adapt and generalize to unfamiliar environments. The core of adaptation and generalization is to transfer the knowledge learned during training to new task scenarios. Meta-RL is considered as one of the most effective approaches to facilitate fast adaptation to new tasks for embodied intelligence. The goal of meta-RL is to learn a policy within a given task distribution, which can efficiently adapt to a new task distribution with minimal data acquisition \cite{finn2017model,duan2016rl,stadie2018some}. This goal of meta-RL, i.e., learning for fast adaptation, is well aligned with the learning strategy of humans and embodied AI. Therefore, understanding the mechanism of fast adaptation in meta-RL agents will shed light on the human and embodied AI.

Task representation, as a critical component of meta-RL, impacts the agent's generalization capabilities, particularly in complex control and navigation tasks. Agents are influenced by different task representations during training \cite{zhang2020learning,yuan2022robust,sodhani2021multi,wang2023meta} and need to effectively learn representations to abstract the shared structures in the task distribution. The ability of task representation of an agent mainly depends on data sampling, such as utilizing data relevant to task generalization. Therefore, how to effectively sample task-relevant data during training is crucial, especially for training off-policy meta-RL agents within an off-policy framework. 
Existing meta-RL algorithms can be categorized into the policy gradient methods \cite{finn2017model,zintgraf2019fast,raghu2019rapid} and context-based methods \cite{wang2016learning,duan2016rl,rakelly2019efficient,zintgraf2019varibad}. These meta-RL methods can be categorized into on-policy and off-policy based on the relationship between the policy employed during the learning process and the policy used to generate the data. Compared to on-policy meta-RL algorithms, the off-policy algorithms are more sample-efficient and also rely more on suitable data sampling strategies~\cite{rakelly2019efficient,rusu2018meta,dorfman2021offline}. However, how the data sampling strategy affects the off-policy meta-RL agents is still unclear.


\begin{figure}[!bt]
\centering
\includegraphics[width=0.7\textwidth]{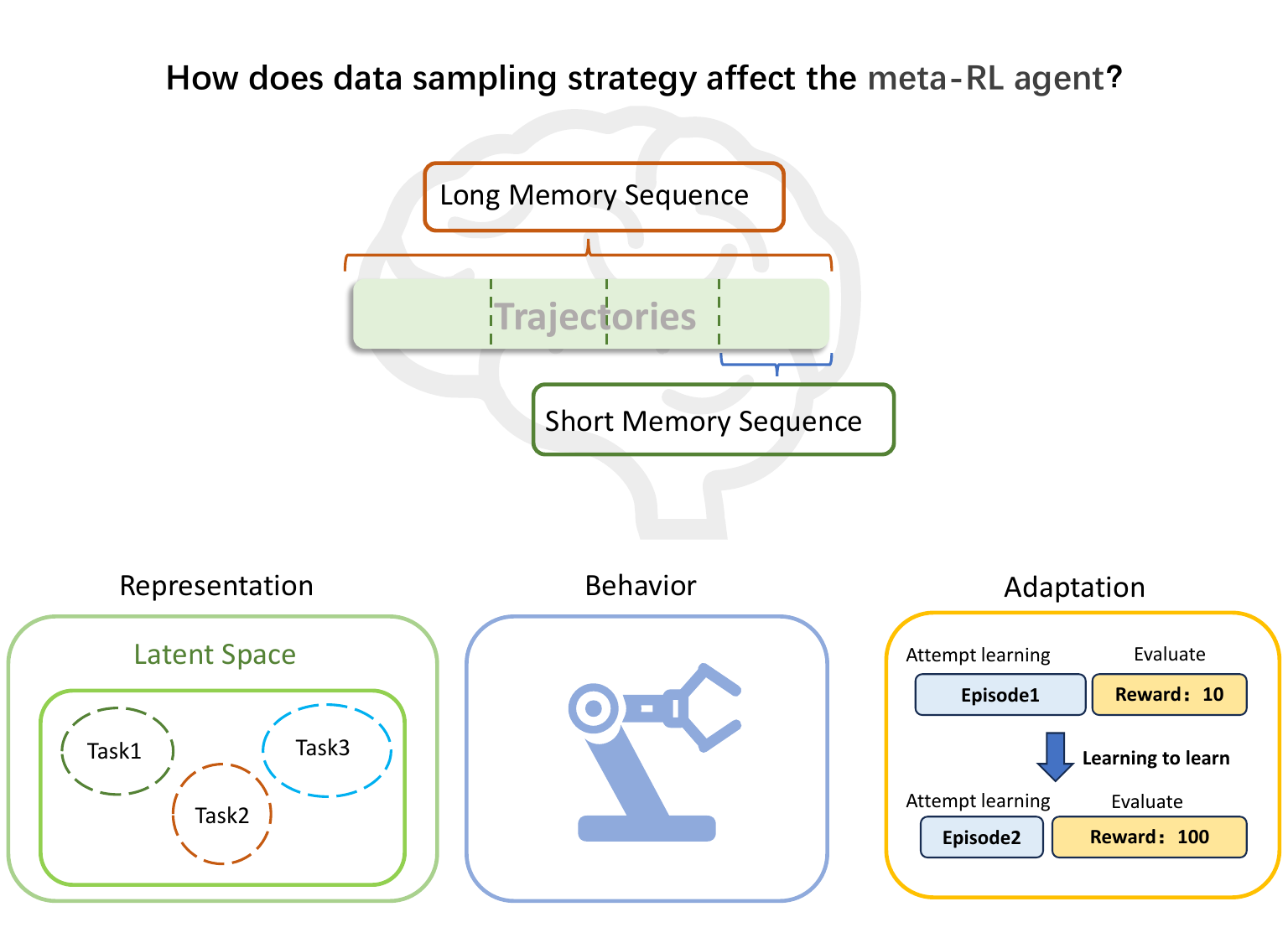}
\caption{Motivations of our work.} \label{fig1}
\end{figure}

In this study, we aim to understand how the data sampling strategy affects the online off-policy meta-RL algorithms, in terms of task representation, agent behaviors and adaptation ability (Figure~\ref{fig1}). To this end, we analyze the exploration capabilities of two types of off-policy meta-RL algorithms based on different data sampling strategies, specifically those based on Thompson sampling \cite{thompson1933likelihood} in PEARL \cite{rakelly2019efficient} and Bayes-optimal policy \cite{duff2002optimal} in VariBAD \cite{zintgraf2019varibad}. By conducting experiments with two meta-RL algorithms based on Bayes-optimal policy and Thompson sampling using different data sampling strategies, we examine the influence of two data sampling strategies, i.e., long and short memory sequence, on task representation, agent behavior, and adaptability of off-policy meta-RL agents through experiments in continuous control tasks in MuJoCo [25] and complex navigation tasks.

Our findings are summarized as follows.
\begin{itemize}
    \item Meta-RL based on Bayes-optimal policy has superior robustness to data sampling distributions compared to Thompson sampling-based Meta-RL method in sparse reward tasks. This robustness originates from the better representation of the unknown environment's dynamics and reward models (Figure~\ref{fig5}, ~\ref{fig6} and~\ref{fig7}).
    \item Experiments on complex robotic navigation tasks demonstrate that although the short memory sampling strategy enables PEARL to converge faster, it does not improve the agent's adaptability; in contrast, the relatively robust off-policy VariBAD algorithm exhibits stronger adaptability (Figure~\ref{fig8}, ~\ref{fig9} and~\ref{fig10}).
    \item The robustness of algorithms to short memory sequence or long memory sequence sampling strategy is associated with their adaptability capabilities.
\end{itemize}


\section{Background and Related Work}

In this section, we primarily introduce the foundational concepts of POMDPs and meta-RL, and related work on task representation in reinforcement learning.

\subsection{POMDP and Meta-RL}


A partially observable Markov decision process (POMDP) \cite{kaelbling1998planning} framework offers a robust mathematical model for decision-making where agents must act under conditions of uncertainty and partial information. A POMDP is defined as a tuple $(S, A, O, T, Z, R, \gamma)$, where $S$ is the state space, containing all possible states of the environment; $A$ is the action space, containing all actions that the agent can perform; $T: S \times A \rightarrow \mathcal{P}(S)$ is the state transition function; $Z: S \times A \rightarrow \mathcal{P}(O)$ is the observation function, defining the probability distribution of generating observations given the next state and action; $R: S \times A \times S \rightarrow \mathbb{R}$ is the reward function, which computes the immediate reward based on the current state, chosen action, and resulting state, and $\gamma$ is the discount factor. The optimization objective of a POMDP is to find a policy $\pi: \mathcal{H} \rightarrow A$, where $\mathcal{H}$ represents all possible sequences of historical information. In off-policy meta-RL methods, sufficient task representation from historical trajectories influences the online performance.

The adaptation process of meta-reinforcement learning agents in unknown environments can be seen as a generalization process within POMDPs with a similar distribution. In conventional reinforcement learning algorithms, policy $\pi$ aims to maximize the expected discounted cumulative reward, expressed as:
\begin{equation}
\mathcal{J}^\pi=\mathbb{E}\left[\sum_{t=0}^{\infty} \gamma^t R\left(s_t, a_t, s_{t+1}\right) \mid \pi\right] . 
\end{equation}

Meta-RL extends the foundational concepts of reinforcement learning by enabling agents to learn how to learn across a variety of tasks, rather than optimizing for a single task. This approach leverages past experience to rapidly adapt to new environments or tasks with minimal additional data. The framework of meta-RL is built around the idea that the skills acquired in previous tasks can inform the agent's policy on unseen tasks, thus reducing the time and data required for learning new tasks.
The objective is to train a learning algorithm that can quickly adapt to new tasks using only a few interactions:
\begin{equation}
\max_{\theta} \mathbb{E}_{\tau \sim p(\tau|\theta, T)} \left[ \sum_{t=0}^T \gamma^t r_t \right],
\end{equation}
where $\theta$ represents the meta-parameters of the policy, and $T$ represents a task sampled from a distribution of tasks. In meta-RL, we aim to perform well across a variety of such tasks.

\subsection{Task Representation in Reinforcement Learning}

Effective task representation involves capturing the essential features of different tasks in a manner that accentuates their commonalities and differences, thus enabling the agent to adapt learned strategies to new, yet similar, scenarios. An expressive representation that captures task variations is vital for reducing the number of interactions needed to adapt to new tasks \cite{humplik2019meta}. Previous work has employed reconstruction loss to train auto-encoders to generate low-dimensional representations of tasks, which are then used to assist in policy learning \cite{lange2010deep,watter2015embed,pere2018unsupervised}. This method is also applicable to the extraction of common features in multi-task settings, often utilized in multi-task reinforcement learning \cite{zhang2021survey,zhang2020learning}. Additionally, some works have used contrastive learning in the latent representation space to obtain robust representations across multiple tasks \cite{laskin2020curl,wang2023meta,yuan2022robust}.

In this paper, we utilize two of the most fundamental and effective meta-RL models. PEARL \cite{rakelly2019efficient} employs an RNN encoder as either the task representation module or the inference module. VariBAD \cite{zintgraf2019varibad,dorfman2021offline} uses a Variational Autoencoder (VAE) \cite{kingma2013auto} as both the representation and prediction module, which not only extracts task representations but also predicts the environmental model during training.

\section{Models and Data Sampling Strategy}

In this section we introduce models used and show how to use long-term memory replay and short-term memory reply in two context-based meta-RL methods.

\subsection{Thompson Sampling and PEARL}


Thompson sampling \cite{thompson1933likelihood} is a Bayesian approach to addressing the exploration-exploitation dilemma and has been effectively applied in the context of meta-RL \cite{ortega2019meta}. In sequential decision-making, meta-learning can be divided into two phases: the first phase involves abstracting a representation of the distribution of training tasks, while the second phase allows the policy to leverage the prior knowledge of the task distribution acquired in the first phase to predict unknown task distributions, achieving rapid adaptation with minimal interactions with the environment. These models are typically comprised of two components: a task inference module and a policy module \cite{beck2023survey,rakelly2019efficient,zintgraf2019varibad,raileanu2020fast}. Thompson sampling can be described as the process of sampling actions from the policy based on posterior predictions \cite{ortega2010minimum}. According to Bayes' theorem, given the historical trajectory \(\tau\), we can update the posterior distribution of \(\theta\): $P(\theta \mid \tau) \propto P(\tau \mid \theta) P(\theta)$, where 
\begin{equation}
    P(\tau \mid \theta)=\prod_{t=0}^{T-1} P\left(s_{t+1} \mid s_t, a_t, \theta\right) P\left(r_t \mid s_t, a_t, \theta\right) .
\end{equation}

We employ PEARL as a meta-RL model based on Thompson sampling on historical information. The inference network (task encoder) \(q_\phi\) encodes the agent's historical information to capture task-relevant sufficient statistics, and $\mathbf{z}$ is the latent embedding of the encoder. During the meta-training phase, the parameters within \(q_\phi\) are optimized by modeling the Q-value function \(Q_\theta(\mathbf{s}, \mathbf{a}, \mathbf{z})\) within the Soft Actor-Critic (SAC) \cite{haarnoja2018soft} algorithm and constrained by the variational lower bound to enhance the inference module's ability to learn information pertinent to the current task being performed, the inference module and policy module use different experience.

\begin{figure}
\includegraphics[width=\textwidth]{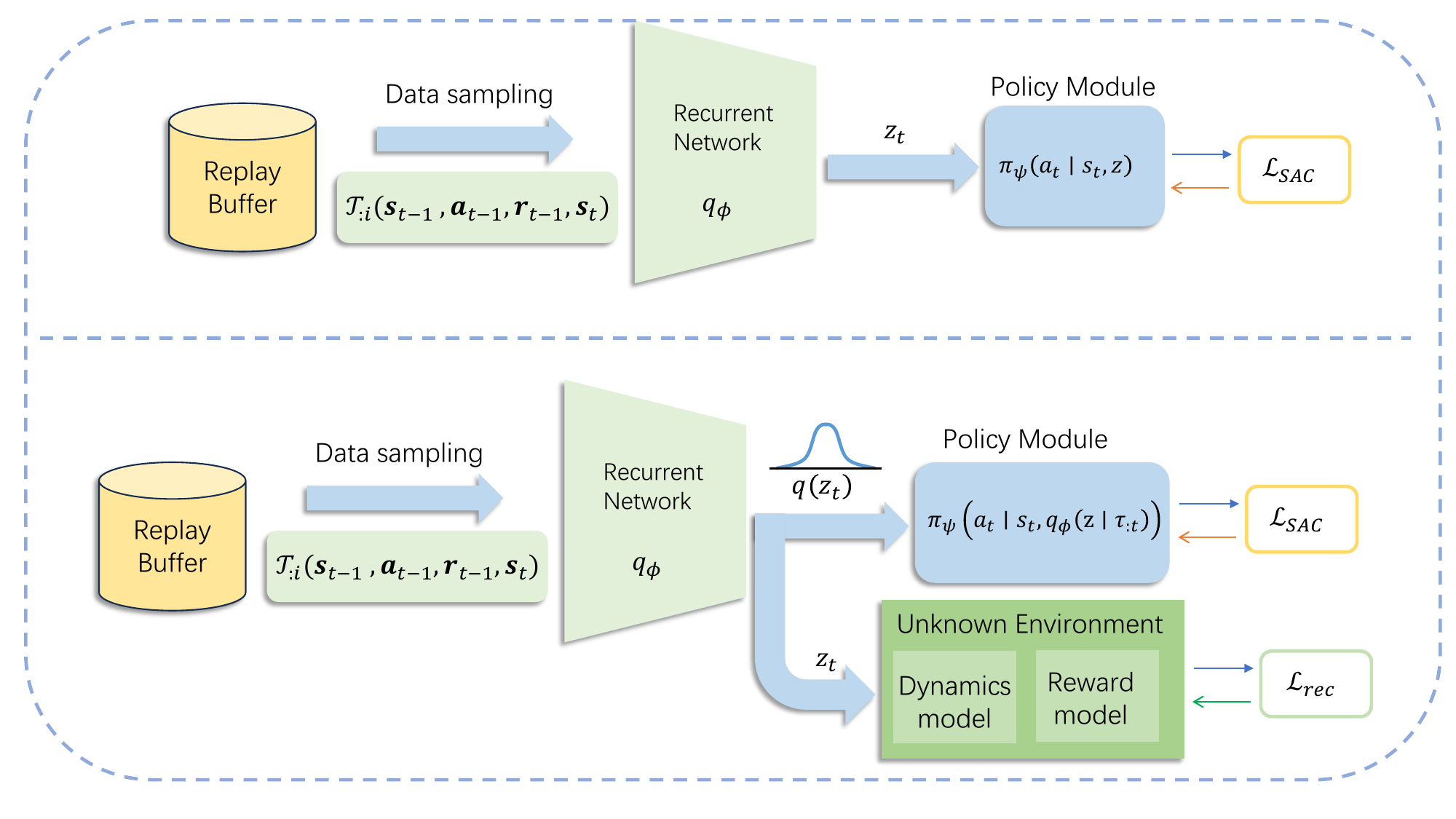}
\caption{PEARL and off-policy VariBAD famework.} \label{fig2}
\end{figure}

In the meta-testing phase, as the agent tackles unknown tasks, it updates the posterior of task history in a manner similar to Thompson sampling through the inference network, enhancing its capability to explore unknown tasks. The structure of PEARL is depicted in the upper part of Figure~\ref{fig2}.

\subsection{Bayes-optimality and VariBAD}

Bayes-optimality is a principle within the broader Bayesian decision theory that dictates selecting actions based on maximizing expected utility, considering all possible outcomes weighted by their probabilities. In the context of meta-RL, a Bayes-optimal policy aims to maximize the expected reward across a distribution of tasks by leveraging a posterior distribution over tasks \cite{duff2002optimal}. This approach inherently balances exploration and exploitation by considering the uncertainty in the environment's dynamics and the reward function. By integrating prior knowledge with observations gathered during interactions with the environment, the Bayes-optimal approach adapts its strategy to better respond to unseen dynamics. In BAMDP, we aim to maximize the expected reward over $T$ time steps:
\begin{equation}
\mathcal{J}(\pi)=\mathbb{E}_{b_0, \pi}\left[\sum_{t=0}^{T-1} \mathbb{E}_{R \sim b_t}\left[R\left(s_t, a_t\right)\right]\right] ,
\end{equation}
where $b_t=p\left(r, p \mid \tau_{: t}\right)$ represents the belief about the current environment dynamics and reward function based on the historical trajectory.

VariBAD implements Bayesian optimal policies under the framework of Bayes-Adaptive MDP (BAMDP) \cite{duff2002optimal,ghavamzadeh2015bayesian} and it is designed to learn a latent representation of the environment’s dynamics, rewards, and task-specific parameters through a variational autoencoder architecture (lower part of Figure~\ref{fig2}). During training, the agent learns a universal policy that is conditioned on both the current state and a latent variable that encodes task-specific information. This latent variable is updated as new data is collected, effectively allowing the agent to infer the underlying task dynamics and adapt its policy accordingly. VariBAD thus enables an agent to perform robustly across a variety of tasks by efficiently learning and updating its understanding of the task environment.

\subsection{Long and Short Memory Sequence Sampling}

\begin{figure}
\includegraphics[width=\textwidth]{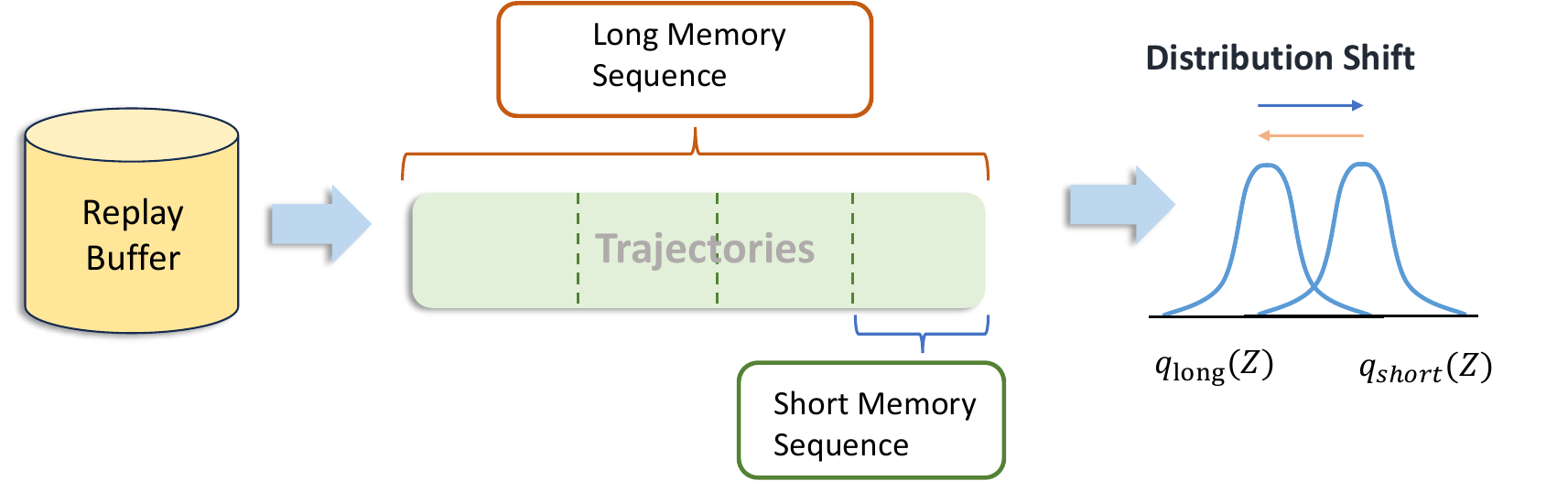}
\caption{Long memory sequence sampling and short memory sequence sampling. Different context sampling strategies can lead to shifts in the distribution of task representations, thereby affecting the agent's exploration and adaptation capabilities.} \label{fig3}
\end{figure}

In this paper, we investigate the impact of data sampling strategies on the prior information extracted by the representation module in off-policy meta-RL, leading to varying exploration outcomes by the algorithm. In our experiments, we have set up long and short memory sequence sampling strategies (Figure~\ref{fig3}). The long memory strategy involves storing all historical information from interactions with the environment in the replay buffer, from which the agent and inference network sample during training. In contrast, the short memory sequence sampling setting requires clearing the replay buffer at the start of each training iteration, ensuring that the agent and inference network can only sample the most recent historical data. Typically, Thompson sampling updates the posterior based on the most recent historical information. In our experiments described in Section 4, we observe that the long memory sequence can significantly disrupt the exploration effectiveness of PEARL, while the off-policy VariBAD remains robust under both settings. The exploration and exploitation trade-off is also reflected in the varying performance of the different representation modules.

\section{Experiments}



In our study, we conduct comparisons of the performance of PEARL and off-policy VariBAD under various data sampling strategies in the continuous control task Sparse Half-Cheetah-Vel within the MuJoCo environment and two challenging navigation tasks, Ant-Semi-Circle and Sparse-Point-Robot. We analyze the robustness of the algorithms and their task adaptation capabilities based on their performance.

\subsection{Task setting}

\subsubsection{Sparse Half-Cheetah-Vel.}

We have modified the original Half-Cheetah environment in MuJoCo to adopt a sparse reward function, where the cheetah receives a reward of +1 only if it moves a distance greater than a specified threshold in each time step; otherwise, the reward is 0.
$$
r_t^{sparse}= \begin{cases}1, & \left\|x_t-x_{\text {goal }}\right\|_2 \leq r \\ 0, & \text {otherwise}.\end{cases}
$$

Subsequently, we utilize the commonly used Half-Cheetah-Vel task in meta-RL, where the objective for the agent is to reach a target velocity as quickly as possible. The target velocity is randomly sampled from a range of 0.0 to 3.0. Consequently, the reward in this environment is structured as follows:
$$
r_t=-\left|v_t-v_{\text {goal }}\right|-0.05 \cdot\left\|a_t\right\|_2^2 + r_t^{sparse} ,
$$
We set up 100 training tasks and 20 testing tasks.

\subsubsection{Sparse-Point-Robot.}


We established the Sparse-Point-Robot environment to evaluate the performance of algorithms in a sparse reward navigation setting. Different goals are set on a semi-circle, with their locations unknown. At the start of each episode, the robot's initial position is randomly placed outside of the semi-circle. The objective is for the robot to locate the target within a single episode. The reward structure is configured as follows:

$$
r = \begin{cases} 
1, & \text{if } r \geq -\text{goal\_radius} \\
0, & \text{otherwise}.
\end{cases}
$$

\subsubsection{Ant-Semi-Circle.}

In the Ant-Semi-Circle task \cite{dorfman2021offline}, an ant robot is required to navigate toward a goal that is randomly positioned on a semi-circle. Unlike the point robot scenario, this task employs the Ant model from the MuJoCo simulation, which introduces increased control complexity. The reward structure is set as follows:
$$
r_t=-\left\|x_t-x_{\text {goal }}\right\|_1-0.1 \cdot\left\|a_t\right\|_2^2 .
$$

\subsection{Convergence of Algorithms}


We conducted experiments in the aforementioned three environments, initially testing the convergence of the algorithms on different tasks under default parameters (PEARL using short sequence, off-policy VariBAD using long  sequence).

\begin{figure}
\centering
\includegraphics[width=\textwidth]{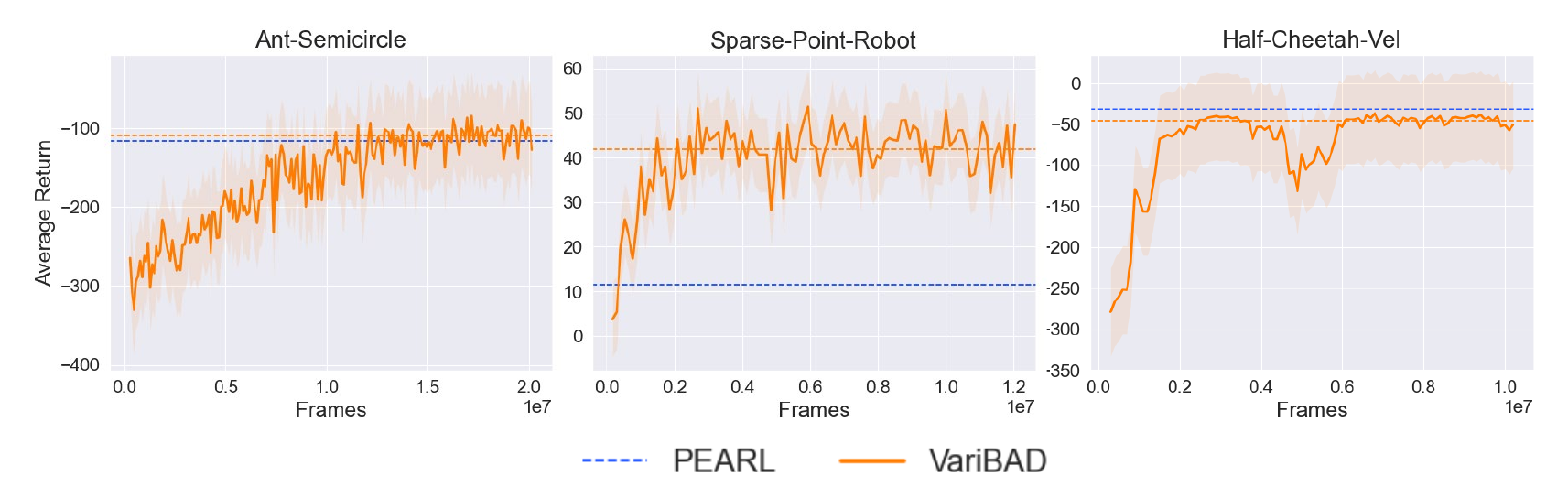}
\caption{Tasks training. Dashed lines correspond to the maximum return achieved by PEARL after 1e6 steps. Solid lines correspond to average return achieved by VariBAD. In the Ant-Semicircle and Half-Cheetah-Vel tasks, PEARL and VariBAD converge to similar average returns. However, in the Sparse-Point-Robot task, VariBAD significantly outperforms PEARL.} \label{fig4}
\end{figure}

\subsection{Robustness of Algorithms}
\begin{figure}
\includegraphics[width=\textwidth]{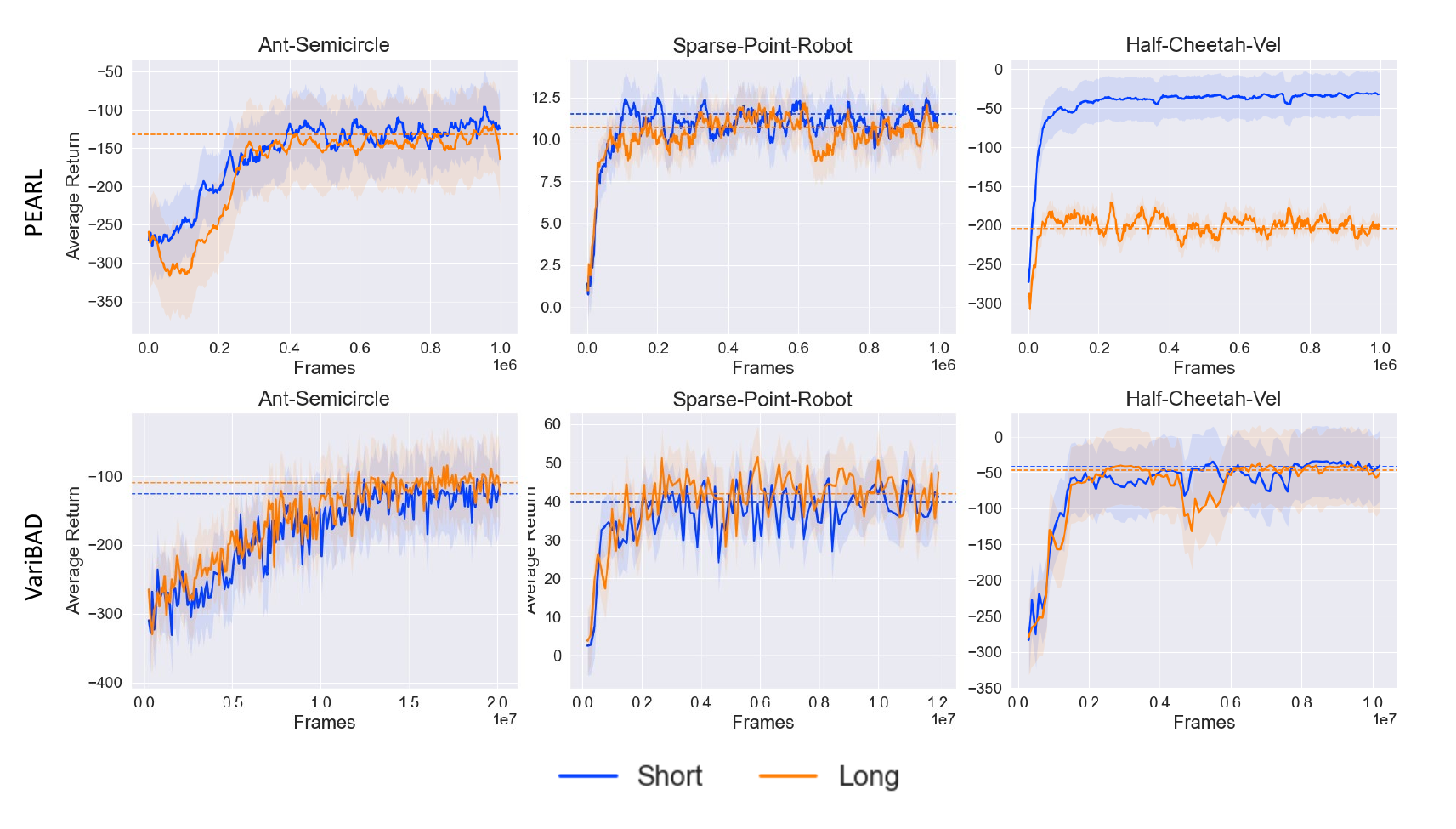}
\caption{The average return during the meta-training phase of PEARL and off-policy VariBAD after using short and long memory sampling strategies. } \label{fig5}
\end{figure}

For algorithm robustness, the analysis mainly focuses on the task representations during the meta-training and meta-testing phases. During the meta-training phase, for tasks involving the control of simulated robots such as Ant-SemiCircle and Half-Cheetah-Vel, PEARL is more susceptible to the influence of memory sequence sampling. Here, short memory sequence sampling proves more advantageous for PEARL's adaptation to new environments, while off-policy VariBAD remains relatively stable in comparison (Figure~\ref{fig5}).

\begin{figure}
\includegraphics[width=\textwidth]{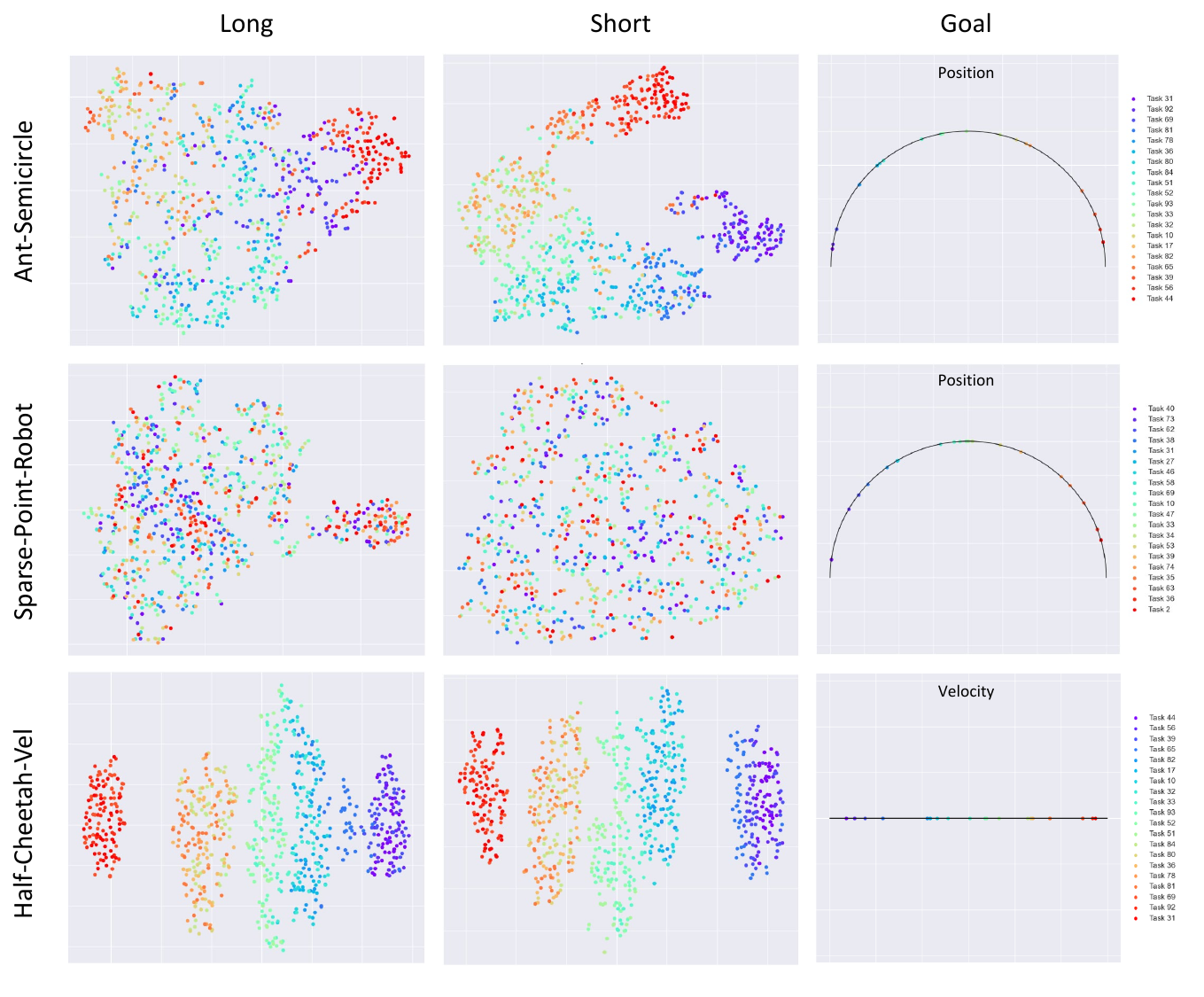}
\caption{The t-SNE visualization of latent embedding during PEARL adaptation to the environment.} \label{fig6}
\end{figure}

\begin{figure}
\includegraphics[width=\textwidth]{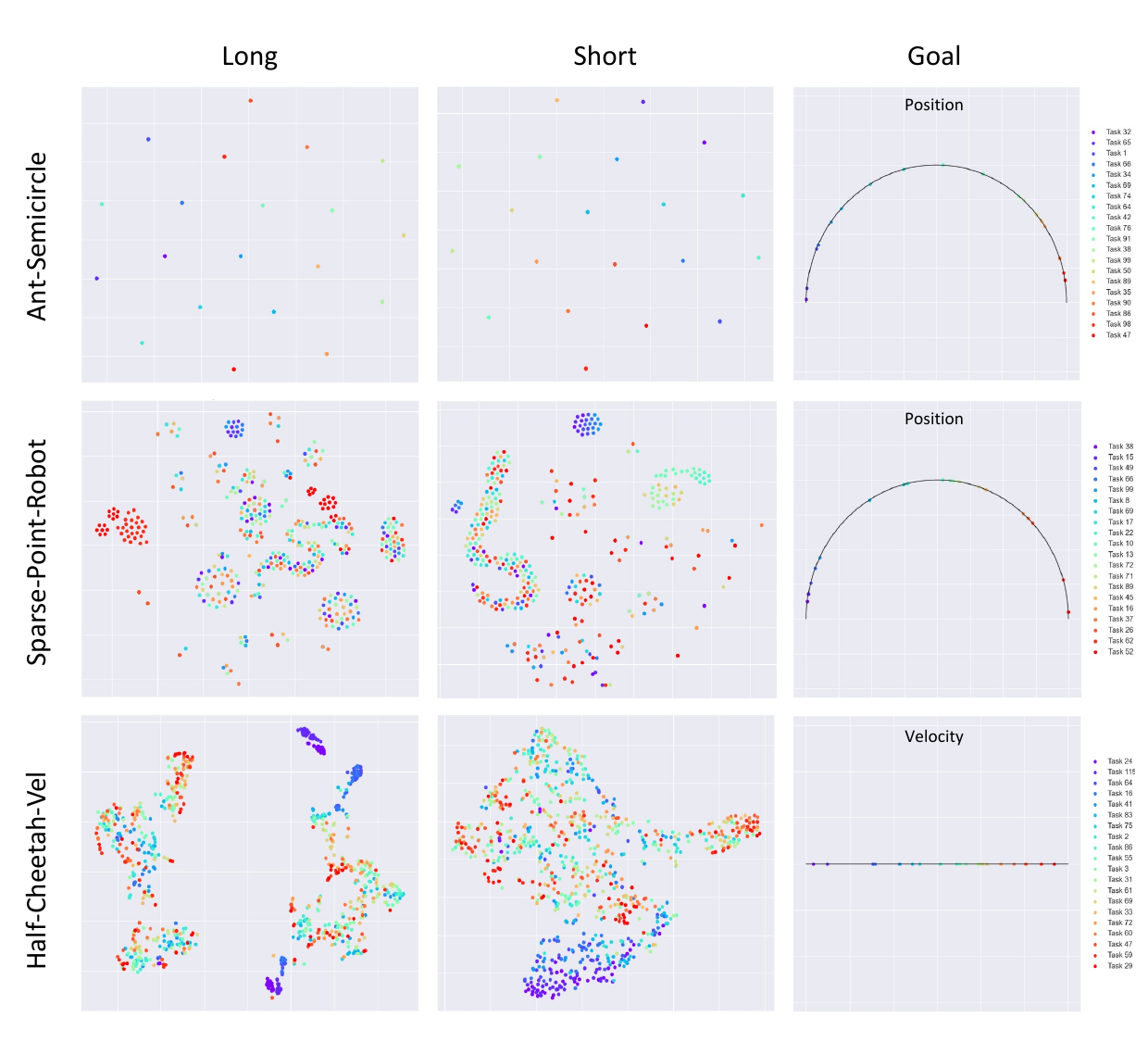}
\caption{The t-SNE visualization of latent embedding during off-policy VariBAD adaptation to the environment.} \label{fig7}
\end{figure}

We randomly generate 20 goals from each environment, and for each goal, the agent performs 40 runs, each containing 5 episodes. We utilize t-SNE \cite{van2008visualizing} to visualize the latent embeddings of the task inference module at the last time step of the fifth episode during the meta-testing phase for PEARL (Figure~\ref{fig6}) and off-policy VariBAD (Figure~\ref{fig7}). For navigation tasks, especially the Sparse-Point-Robot task, PEARL does not sufficiently learn the task-relevant information of the environment, whereas off-policy VariBAD exhibits better performance, forming clusters for similar positions on the semicircle. In the case of Ant-Semi-Circle, off-policy VariBAD accurately represents each target's position on the semicircle in the latent space, whereas the clusters formed by PEARL are more dispersed. In the Half-Cheetah-Vel environment, the representation of off-policy VariBAD is more sensitive to the memory sequence sampling strategy because, inherently, based on Bayes-optimality, it requires sampling of the whole history, and the representation in continuous control tasks is significantly influenced by the history.

 

\renewcommand{\floatpagefraction}{.9}

\subsection{Tasks Adaptation Performance}
\begin{figure}
\includegraphics[width=\textwidth]{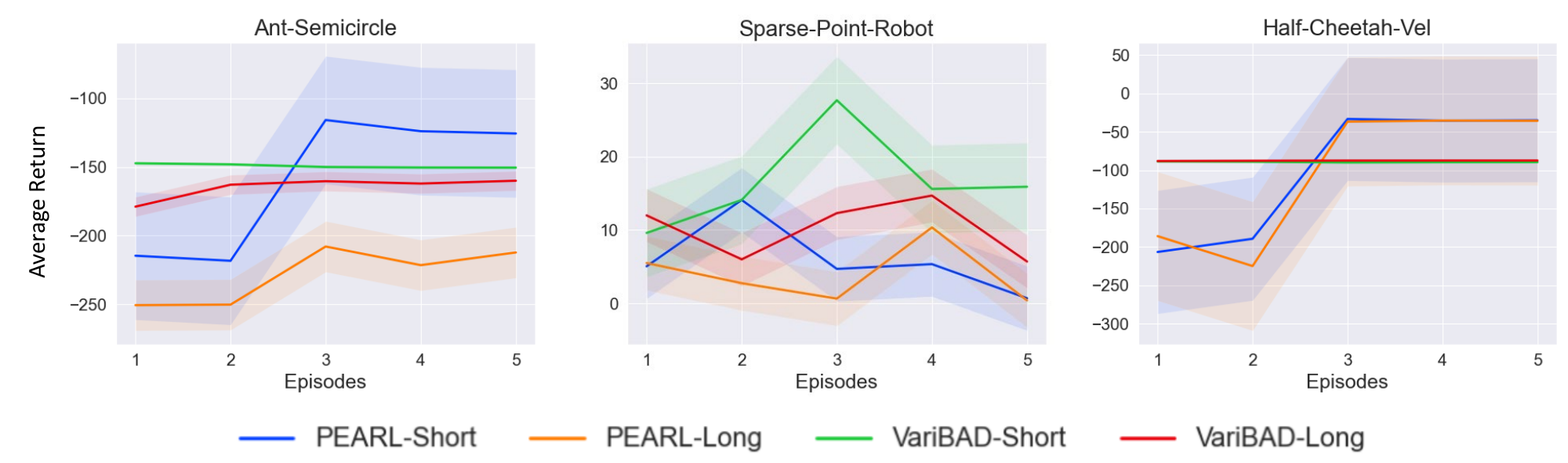}
\caption{Adaptation performance of PEARL and off-policy VariBAD using short and long memory sequence sampling strategy.} \label{fig8}
\end{figure}

To compare the adaptability of PEARL and VariBAD using different sampling strategies in unknown environments, we conducted rollouts of 5 episodes in each environment (Figure~\ref{fig8}). The experiments revealed that the long memory sequence sampling strategy prevented PEARL from adapting effectively to navigation tasks and from achieving satisfactory results in the Sparse-Point-Robot task, consistent with its performance during the meta-training phase. In contrast, off-policy VariBAD demonstrated stable adaptability, successfully adjusting to tasks and achieving high average returns from the first episode across all three environments. Moreover, VariBAD showed less sensitivity to memory sequence sampling strategy, with agents trained using short memory sequence in the Sparse-Point-Robot task exhibiting enhanced exploratory capabilities.

\begin{figure}
\includegraphics[width=\textwidth]{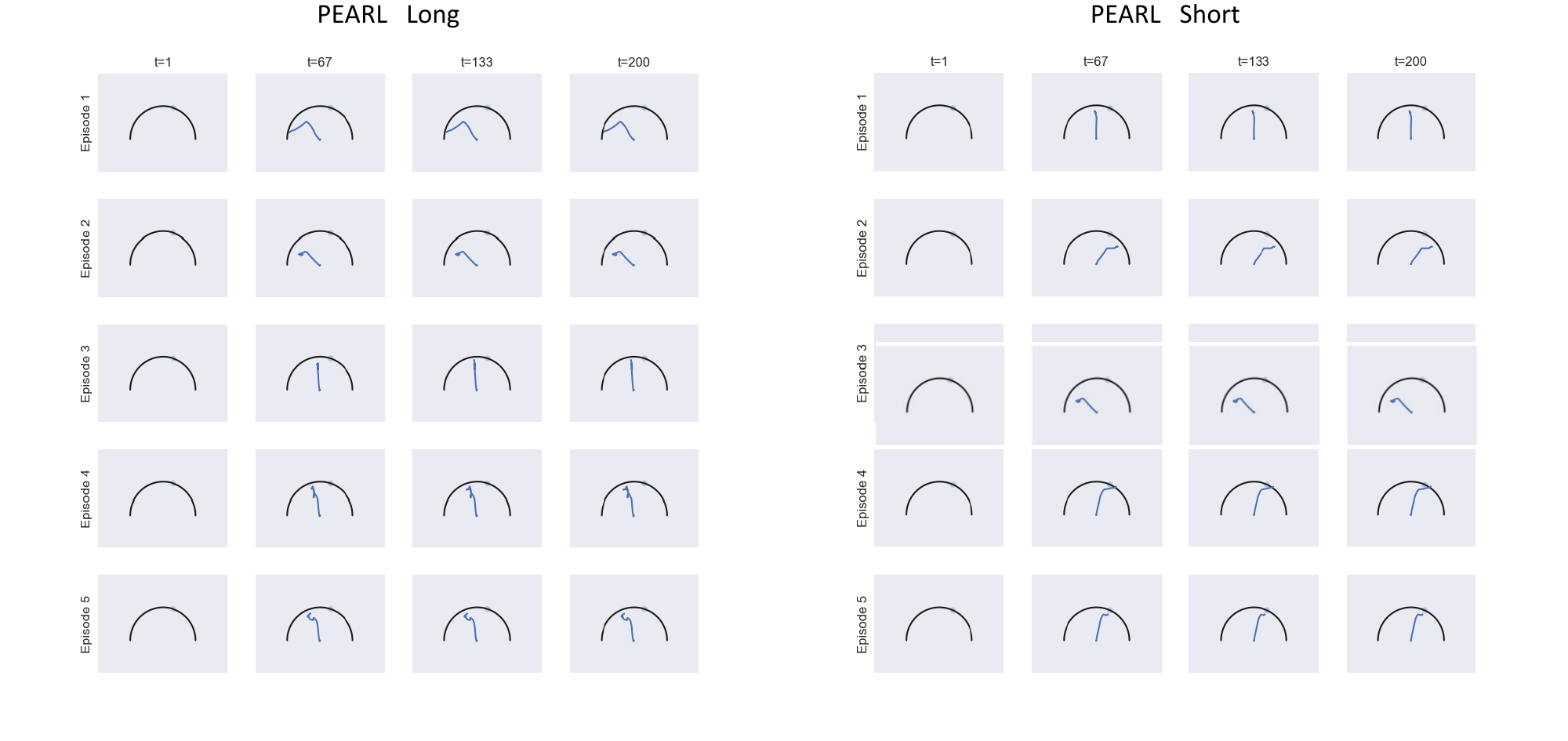}
\caption{Behavior visualization of PEARL during adaptation in Ant-Semi-Circle.} \label{fig9}
\end{figure}

\begin{figure}
\includegraphics[width=\textwidth]{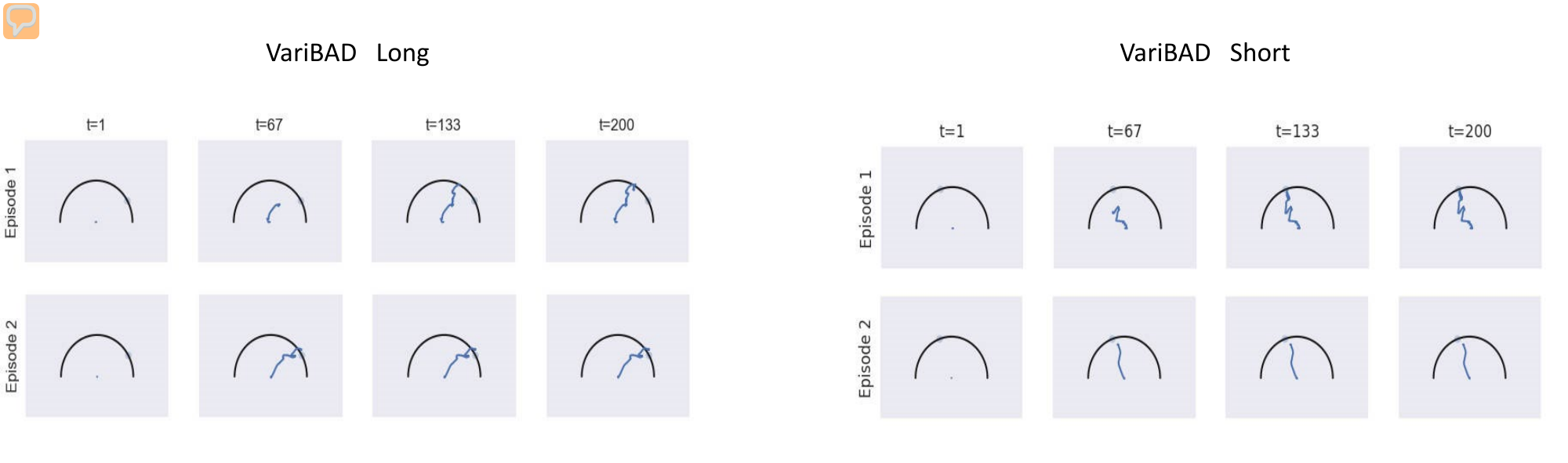}
\caption{Behavior visualization of off-policy VariBAD during adaptation in Ant-Semi-Circle.} \label{fig10}
\end{figure}

Furthermore, to intuitively assess the impact of sampling strategies on the exploratory capabilities of agents during the meta-testing process, we visualized the exploration trajectories of agents in the Ant-Semi-Circle environment. The length of the memory sequence affects PEARL's exploratory capabilities, primarily because Thompson sampling focuses on updating the encoder's posterior based on recent context, hence performing better with a short memory sequence length. Figure~\ref{fig9} shows that PEARL, when using short memory, can successfully reach the target in all five episodes; however, when switched to long memory, the agent fails to accurately locate the target for the same test task. Figure~\ref{fig10} demonstrates that off-policy VariBAD can successfully find the target in two episodes and is unaffected by the memory sequence length. This indicates that applying Bayes-optimality to off-policy meta-reinforcement learning results in stronger online adaptability to complex environments and more robust task representation compared to algorithms based on Thompson sampling.

\renewcommand{\floatpagefraction}{.9}

\section{Conclusion}

\subsubsection{Discussion.}

In this work, we comprehensively compared the impact of memory sequence length sampling on task representation, agent behavior, and the exploration and adaptation capabilities of two types of context-based off-policy meta-RL algorithms. The off-policy VariBAD algorithm, based on Bayes-optimality, demonstrated stronger robustness in sparse reward environments and its adaptability to unknown environments and task representations were less influenced by the training-time memory sequence. Although the PEARL algorithm, based on Thompson sampling, achieved similar average returns during the training phase, different memory lengths significantly affected its exploratory capabilities during the adaptation phase, fundamentally because the memory sequence length can cause shifts in the distribution of task representations.

\subsubsection{Future work.}

Task representation extraction, especially in multi-task scenarios, is crucial for an agent's ability to adapt to unknown tasks \cite{sodhani2021multi,wang2021improving,zhang2020learning,lee2021improving}. This paper explores the impact of memory sequence length on shifts in task feature distributions, providing insights into how to maximize the extraction of information relevant to the current task from limited historical data and generate a robust representation of the task distribution. This prompts us to employ effective representation learning methods in our subsequent work to enhance the performance of the task inference module.

\section*{Acknowledgments}

This work was funded in part by the National Key R\&D Program of China (2021YFF1200804), Shenzhen Science and Technology Innovation Committee (2022410129, KCXFZ20201221173400001, KJZD20230923115221044).

\end{document}